\documentclass[journal]{IEEEtran}

\usepackage{graphicx}
\usepackage{amsmath}
\usepackage{amssymb}
\usepackage{xcolor}
\usepackage{verbatim}
\usepackage{hyperref}
\usepackage{algorithm}
\usepackage{algpseudocode}
\usepackage{listings}
\usepackage{caption}
\usepackage{subcaption}
\usepackage{authblk}
\usepackage{multirow}
\usepackage{amsthm}
\usepackage{amsmath}
\usepackage{multicol}
\usepackage{multirow}
\usepackage{makecell}
\usepackage{yfonts}
\usepackage{booktabs}
\usepackage[noadjust]{cite}
\usepackage[labelformat=simple]{subcaption}

\def\x{\mathbf{x}}
\def\f{\mathbf{f}}

\def\x{\mathbf{x}}

\newtheorem{theorem}{Theorem}

\newtheorem{definition}{Definition}

\newtheorem{prop}{Proposition}

\begin{document}

\title{Cloud-RAIN: Point Cloud Analysis with Reflectional Invariance}

\author{
Yiming Cui$^{1}$,
Lecheng Ruan$^{2}$,
Hang-Cheng Dong$^{3}$,
Qiang Li$^4$,
Zhongming Wu$^5$,
Tieyong Zeng$^5$, 
Feng-Lei Fan$^{5*}$
\thanks{$^*$Feng-Lei Fan (hitfanfenglei@gmail.com) is the corresponding author.}
\thanks{
$^1$Yiming Cui (cuiyiming@ufl.edu) is with Department of Electrical and Computer Engineering, University of Florida, Gainesville, FL, USA.}
\thanks{$^2$Lecheng Ruan is with BIGAI, Beijing, China.}
\thanks{$^3$Hang-Cheng Dong is with School of Instrumentation Science and Engineering, Harbin Institute of Technology, Harbin, China.}
\thanks{$^4$Qiang Li is with Department of Electrical and Computer Engineering, McMaster University, Hamilton, ON, Canada. }
\thanks{$^5$Zhongming Wu, Tieyong Zeng, and Feng-Lei Fan are with Department of Mathematics, The Chinese University of Hong Kong, N. T. Hong Kong.}}

\maketitle

\begin{abstract}
The networks for point cloud tasks are expected to be invariant when the point clouds are affinely transformed such as rotation and reflection. So far, relative to the rotational invariance that has been attracting major research attention in the past years, the reflection invariance is little addressed. Notwithstanding, reflection symmetry can find itself in very common and important scenarios, \textit{e.g.}, static reflection symmetry of structured streets, dynamic reflection symmetry from bidirectional motion of moving objects (such as pedestrians), and left- and right-hand traffic practices in different countries. To the best of our knowledge, unfortunately, no reflection-invariant network has been reported in point cloud analysis till now. To fill this gap, we propose a framework by using quadratic neurons and PCA canonical representation, referred to as Cloud-RAIN, to endow point \underline{Cloud} models with \underline{R}eflection\underline{A}l \underline{IN}variance. We prove a theorem to explain why Cloud-RAIN can enjoy reflection symmetry. Furthermore, extensive experiments also corroborate the reflection property of the proposed Cloud-RAIN and show that Cloud-RAIN is superior to data augmentation. Our code is available at \url{https://github.com/YimingCuiCuiCui/Cloud-RAIN}.
\end{abstract}

%
\IEEEpeerreviewmaketitle
        \vspace{-0.3cm}
        
\section{Introduction}
Recently, point clouds have been intensively studied due to their broad applications as a basic representation of 3D models in autonomous driving and robotics. Deep learning techniques are introduced into this field for either direct analysis \cite{zhou2017voxelnet, 7526450, su15mvcnn,atzmon2018point} or indirect analysis represented by PointNet~\cite{qi2017pointnet} and its successors~\cite{qi2017pointnet++, liu2018flownet3d}. As a point cloud is an unordered set, previous direct analysis models focus on the invariance to sample permutation. However, in addition to the invariance to point permutation, 
the networks are also supposed to infer consistently when the point clouds are affinely transformed such as translation, rotation, and reflection. While the bias caused by translation can be counteracted easily by centralization, past years have witnessed a plethora of works \cite{sun2019srinet,esteves2018learning,zhang2019rotation,li2021rotation} proposed to address the invariance to rotation and achieved the encouraging success, \textit{e.g.,} rotation data augmentation, spherical-related convolutions, explicit local rotation-invariant features, and canonical poses. 
 
\begin{figure}[!bt]
    \centering
    \includegraphics[width=0.9\linewidth]{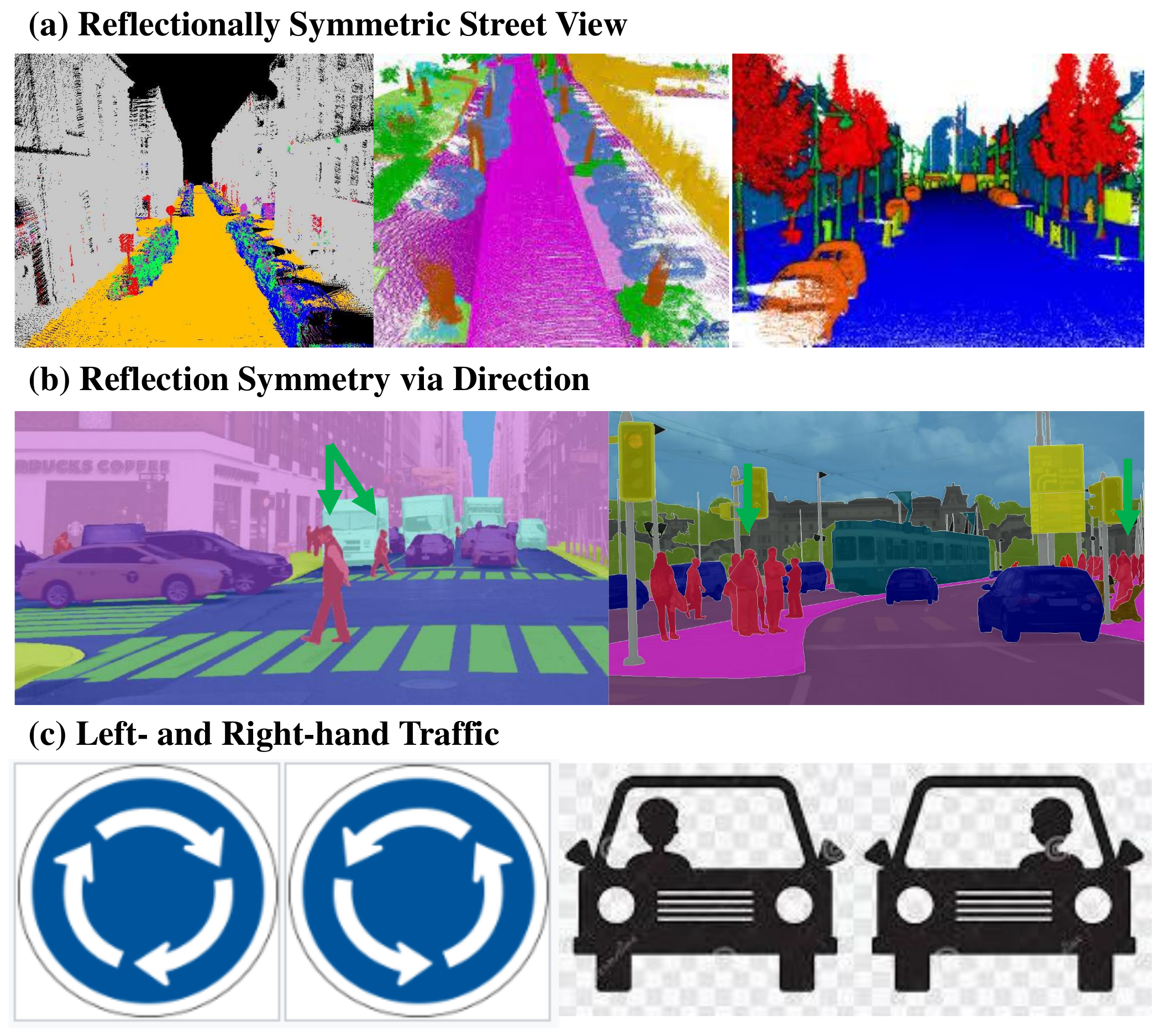}
    \caption{In self-driving scenarios, front view of cars often exhibits reflection symmetry, which could include (a) static reflection symmetry of structured streets, (b) dynamic reflection symmetry from bidirectional motion of moving objects (such as pedestrians), and (c) left- and right-hand traffic practices in different countries (such as UK and US).}
    \label{fig:symmetry}
    \vspace{-0.6cm}
\end{figure}

However, relative to the rotational symmetry, the reflection symmetry is little addressed. Notwithstanding, reflection symmetry can find itself in very common and important scenarios. Often, reflection symmetry appears due to the symmetric city planning. Figure \ref{fig:symmetry}(a) shows one such example in autonomous driving, where the driver's front view exhibits reflection symmetry in well-structured traffic scenes; Also, reflection symmetry is obtained by the direction of movement. Figure \ref{fig:symmetry}(b) manifests that pedestrians crossing the street to the left and to the right are symmetric in reflection. A devastating car accident may happen if a self-driving car only identifies the pedestrians walking along one direction and fails the opposite; In addition, reflectional symmetry can be given rise by the left- and right-hand traffic practices in different countries, as Figure \ref{fig:symmetry}(c) shows. Furthermore, we argue that reflection symmetry is more common than rotation symmetry in some important scenarios such as driving, where the environment around an autonomous vehicle is unlikely to rotate severely but likely to exhibit reflection symmetry. Therefore, reflectional invariance should be examined dedicatedly in a point cloud model.

However, to the best of our knowledge, no reflection-invariant network has been claimed in point cloud analysis till now. To deal with this issue, 
here we shift our attention to the quadratic neurons and networks. Recently, motivated by introducing neuronal diversity in deep learning, the so-called quadratic neuron \cite{fan2018new} was designed by using a simplified quadratic function to replace the inner product in the conventional neuron. Hereafter, we call a network made of quadratic neurons a quadratic network and a network made of conventional neurons a conventional network. \textit{Why can quadratic networks address reflectional symmetry?} Here, we prove that a quadratic network can approximate a reflectionally symmetric function well compared to conventional networks. The idea is that due to the invariance to the sign flipping of the input, the power term in the quadratic neuron can directly deal with the reflection of the input with respect to axis-aligned planes. Next, we combine quadratic networks with PCA to build a point \underline{Cloud} method with \underline{R}eflection\underline{A}l \underline{IN}variance, referred to as Cloud-RAIN. While quadratic neurons deal with the reflectional symmetry due to the sign flipping, the PCA canonical transform derives canonical poses of a point cloud, which can transform the problem of reflection across arbitrary planes into a problem of sign flipping. Thus, Cloud-RAIN can tackle a general arbitrary-plane reflectional invariance problem. Our contributions are threefold: 

\textbullet~We identify reflection invariance as an essential yet overlooked concern in point cloud analysis. 

\textbullet~ We prove a theorem showing that a quadratic network is powerful in approximating a reflectionally-invariant function. Then, we propose Cloud-RAIN that combines quadratic neurons and the PCA transform for point cloud applications.

\textbullet~Empirically, Cloud-RAIN is invariant to reflection and much better than reflection data augmentation.

\vspace{-0.2cm}
\section{Related Works}

 \textbf{Deep Learning-based Point Cloud Analysis.}
Currently, deep learning methods for point cloud analysis can be divided into two classes: indirect and direct. Indirect methods transform point clouds either to multi-view images or volumetric data followed by a 2D CNN or a 3D CNN, including VoxelNet~\cite{zhou2017voxelnet}, SPLATNet~\cite{su18splatnet}, and PCNN~\cite{atzmon2018point}. In contrast, direct methods represented by PointNet~\cite{qi2017pointnet}, PointNet++~\cite{qi2017pointnet++}, RS-CNN~\cite{liu2019rscnn}, PointWeb~\cite{zhao2019pointweb}, KPConv~\cite{thomas2019KPConv}, PointConv~\cite{wu2018pointconv}, and so on~\cite{hu2020randlanet, tang2022contrastive, ma2022rethinking, nie2022pyramid}. Direct methods are able to capture extrinsic features like 3D corners, edges, and local shape parts. 
Concurrently, graph-based methods build a graph by considering each point in a point cloud
as a vertex and generating directed edges based on the vertices' neighborships. Then, graph networks perform feature learning in spatial or spectral domains. Representative graph models include DGCNN~\cite{wang2019dynamic},
LDGCNN~\cite{zhang2019linked}, DeepGCNs\cite{li2019deepgcns}, PointGCN~\cite{ZhangR_18_gcnn_point_cloud}, GAC-Net~\cite{8954040}, and so on. Recently, Transformers \cite{vaswani2017attention} were more and more introduced in the point cloud tasks \cite{zhao2021point,lai2022stratified, park2022fast,zhang2022patchformer}. One common concern regarding deep models is that these models are vulnerable to symmetric transforms when data augmentation is not applied.

\textbf{Rotation Symmetry.}
Intensive investigations have been conducted on the rotation-invariant property. These studies are roughly divided into three categories: i) rotation-invariant operations \cite{poulenard2019effective, rao2019spherical, esteves2018learning,zhang2019rotation,fuchs2020se, deng2021vector}, which designs rotation-invariant convolutions or attentions in a network to achieve invariance; ii) explicit local rotation-invariant features \cite{li2021rotation, sun2019srinet, chen2019clusternet, xu2021sgmnet, yu2020deep}, which handcrafts rotationally invariant features based on the intrinsic geometric relations of point clouds such as relative distances and angles; and iii) canonical poses \cite{fujiwara2020neural, xiao2020endowing, li2021closer}, which transforms
point cloud data to their PCA-based canonical poses that preserve shape information to
achieve rotational invariance. 
We think that reflection symmetry is more important than rotation symmetry in some important applications such as driving, where buildings, views, and pedestrians around a normally-moving car are unlikely to encounter severe rotation but very likely to exhibit reflection symmetry. However, unfortunately, endowing a point cloud model with the reflection-invariant property was little studied. 

\textbf{Polynomial Neural Networks.} The story of polynomial neurons begins with the Group Method of Data Handling (GMDH \cite{ivakhnenko1971polynomial}), which gradually learns  higher-order terms as a feature extractor. Furthermore, the so-called higher-order unit was proposed in \cite{poggio1975optimal,giles1987learning} by adding a nonlinear activation $\sigma(\cdot)$ after GMDH: $y=\sigma(Y(\x_1,\cdots,\x_n))$. 
To balance the power of high-order units and parametric efficiency, Milenkoiv \textit{et al.} \cite{milenkovic1996annealing} only utilized linear and quadratic terms and proposed to use the annealing technique to find optimal parameters.
Recently, higher-order, particularly quadratic units were intensively studied \cite{zoumpourlis2017non, tsapanos2018neurons, chrysos2021deep, fan2018new, xu2022quadralib} from the perspective of neuronal diversity. The neurons in \cite{tsapanos2018neurons, goyal2020improved,bu2021quadratic,xu2022quadralib} are a special case of \cite{fan2019quadratic}. \cite{tsapanos2018neurons} excludes the power term, and \cite{goyal2020improved} only has the power term, which renders an incomplete quadratic representation. This work intends to make the first attempt to introduce quadratic neurons into point cloud analysis to realize reflection invariance.

\vspace{-0.1cm}

\section{Power of Quadratic Networks in Reflectional Invariance}

Here, we illustrate why quadratic neurons are suitable for point cloud tasks requiring reflection invariance. We provide a theorem that a quadratic network is powerful in approximating a symmetric function, \textit{e.g.}, using much fewer parameters than a conventional one. This suggests that for tasks requiring a model to be reflectionally symmetric, quadratic networks are favored over conventional ones, which makes the employment of quadratic networks in reflectional invariance well justified. The core idea of this theorem is that the power term in a quadratic neuron are straightforwardly reflection-invariant, while it is costly for the conventional to learn power terms.

\textbf{Quadratic neurons}. Given an input vector $\f=(f_1,f_2,\cdots,f_n)$, a conventional neuron computes an inner product plus a bias $b$ followed by a nonlinear activation $\sigma$: 
\begin{equation} 
\sigma(\f^\top\mathbf{w}+b),
\label{eqn:linear}
\end{equation}
while a quadratic neuron computes two
inner products and one power term and then
feeds them into a nonlinear activation function. Mathematically, its inner-working is
\begin{equation}
\sigma\big((\f^\top\mathbf{w}_1+b_1)(\f^\top\mathbf{w}_2+b_2)+(\f\circ\f)^\top\mathbf{w}_3+b_3\big),
\label{eqn:qn}
\end{equation}
where $\circ$ is a Hadamard product, $\mathbf{w}_1$, $\mathbf{w}_2$, and $\mathbf{w}_3$ are parameters. Compared to the designs of quadratic neurons in \cite{zoumpourlis2017non, jiang2020nonlinear} with a complexity $\mathcal{O}(n^2)$, Eq. \ref{eqn:qn} is much more compact with the complexity of $\mathcal{O}(3n)$, scaling linearly with the number of features $n$. 

\begin{definition}[reflectionally invariant function] We call a function $h(\x)=h(\x_1,\x_2,\cdots,\x_d): \left(\mathbb{R}^N\right)^d \to \mathbb{R}$ reflectionally invariant, if for any $\mathbf{\pi} \in \{-1,1\}^d$, $h(\x)$ satisfies
\begin{equation}
    h(\x_1,\x_2,\cdots,\x_d) =  h(\mathbf{\pi}_1\x_1,\mathbf{\pi}_2\x_2,\cdots,\mathbf{\pi}_d\x_d),
\end{equation}
where $N$ is the number of samples, $d$ is the number of features, $\x_i \in \mathbb{R}^N$, and $\prod_{i=1}^d \pi_i = -1$.
\end{definition}
When $\prod_{i=1}^d \pi_i = -1$, $\mathbf{\pi}_1\x_1,\mathbf{\pi}_2\x_2,\cdots,\mathbf{\pi}_d\x_d$ is a reflection of $\x_1,\x_2,\cdots,\x_d$. When $\prod_{i=1}^d \pi_i = 1$, $\mathbf{\pi}_1\x_1,\mathbf{\pi}_2\x_2,\cdots,\mathbf{\pi}_d\x_d$ is a rotation of $\x_1,\x_2,\cdots,\x_d$. It is also important for a model to maintain rotation invariance over samples. However, previous literature has studied rotation invariance. Therefore, rotation invariance is not the focus of our draft.  

\begin{prop}[Proposition 2 in \cite{yarotsky2017error}]
Given $M>0$ and $\epsilon \in(0,1)$, there is a ReLU network with two inputs that implements a function  $\widetilde{\times}: \mathbb{R}^{2} \rightarrow \mathbb{R}$  so that
a) for any inputs  $x, y $, if  $|x| \leq M$ and $|y| \leq M$, then  $|\widetilde{\times}(x, y)-x y| \leq \epsilon $;
b) if $ x=0 $ or  $y=0$, then  $\widetilde{\times}(x, y)=0 $;
c) the number of computation units in this ReLU network is $O(\ln(1/\epsilon))$.
\label{product}
\end{prop}

\begin{prop}[Lemma 1 in \cite{fan2023one}]
There exists a one-hidden-layer quadratic ReLU network that can implement a mapping  $\tilde{\times}: \mathbb{R}^{2} \rightarrow \mathbb{R} $, satisfying that i)  $\tilde{\times}(x, y)=x y$ ; ii)  $\tilde{\times}(x, y)$  has 2 quadratic neurons and accordingly 4 parameters.
\end{prop}

\begin{theorem} Let $f: \left(\mathbb{R}^{N}\right)^d \rightarrow \mathbb{R}$ be a continuously differentiable and totally symmetric function, where  $\Omega$ is a compact subset of  $\mathbb{R}^{d} $. Let  $\delta>0$, then there exists $\tilde{\mathbf{F}}: \left(\mathbb{R}^{N}\right)^d \rightarrow \mathbb{R}$, such that for any $X$, we have
\begin{equation}
   \sup_X \left|f(X)-\tilde{\mathbf{F}}(X)\right| \leq \delta,
\end{equation}
where $\tilde{\mathbf{F}}(X)=\sum_{l=1}^L c_l \left( \widetilde{\times}_{i=1}^{N} \widetilde{\times}_{\alpha=1}^{d} X_{i, \alpha}^{\phi_{i, \alpha}(l)} \right)$, and $L$ is a fundtion of $\delta$, $c_l$ is a coefficient for different $l$, and $\phi_{i, \alpha}(l)$ is the power function dependent on $i$ and $\alpha$.
To represent $\tilde{\mathbf{F}}$ in the sense of the max norm, a conventional network needs $NdLO(\ln(1/\delta))$ parameters, while a quadratic network only needs $4NdL$ parameters.
\label{thm:representation}
\end{theorem} 

\textbf{The sketch of proof:} We first show a polynomial approximation scheme to the symmetric function $f$, which is $F(\x)$. Then, we construct the quadratic and the conventional networks to represent $F(\x)$, respectively, with a comparison of the approximation error and the number of parameters.
\begin{proof}
\textbf{Step 1}. We first consider the case of $N=1$. According to the Stone-Weierstrass theorem, a polynomial can approximate any continuous function defined on a closed interval. Similarly, based on the fundamental theorem of symmetric polynomials \cite{macdonald1998symmetric}, $f$ can be naturally approximated by a polynomial of elementary symmetric polynomials made of power terms. Mathematically, for any symmetric polynomial $P$, we have
$P(X)=Q\left(e_{1}(X), \ldots, e_{d}(X)\right)$,
where $Q$ is some polynomial, and $e_{k}$ is as the following:
\begin{equation}
    e_{k}(X)=\sum_{1 \leq j_{1}<j_{2}<\cdots<j_{k} \leq d} x_{j_{1}}^2 x_{j_{2}}^2 \cdots x_{j_{k}}^2.
\end{equation}

To extend the result to the case of $N>1$, the monomial takes the form 
\begin{equation}
F(X)=\prod_{i=1}^{d} \prod_{\alpha=1}^{N} X_{i, \alpha}^{\phi_{i, \alpha}},
\end{equation}
where $\phi_{i, \alpha}$ is an even integer, which ensures that this monomial is a reflectionally invariant function. $\phi_{i, \alpha}$ is determined by the target function and the index $\alpha, i$.
The approximation
using a symmetric polynomial can be a linear combination of $L$ symmetrized
monomials.
\begin{equation}
    \mathbf{F}(X)= \sum_{l=1}^L c_l F_{i}^{(l)} (X)= \sum_{l=1}^L c_l \left(\prod_{i=1}^{d} \prod_{\alpha=1}^{N} X_{i, \alpha}^{\phi_{i, \alpha}(l)}\right),
\end{equation}
where $L$ is a function of $\delta$, such that 
\begin{equation}
   \sup_X \left|f(X)-\mathbf{F}(X)\right| \leq \delta/2.
\end{equation}

\textbf{Step 2}. Now, let us use a conventional and a quadratic neural network to approximate $\mathbf{F}(X)$. 

\underline{Approximation by a conventional network}:
We can approximate this product by a conventional neural network with the help of Proposition \ref{product}.
Let $\widetilde{\times}$ be the approximate multiplication. 
Applying $\widetilde{\times}$ in a compositional manner,

\begin{equation}
\underset{X}{\sup} \left | \widetilde{\times}_{i=1}^{N} \widetilde{\times}_{\alpha=1}^{d} X_{i, \alpha}^{\phi_{i, \alpha}}-\prod_{i=1}^{N} \prod_{\phi=1}^{d} X_{i, \alpha}^{\phi_{i, \alpha}} \right | \leq \delta/2 .
\end{equation}

Let $\tilde{\mathbf{F}}=\sum_{l=1}^L c_l \left( \widetilde{\times}_{i=1}^{N} \widetilde{\times}_{\alpha=1}^{d} X_{i, \alpha}^{\phi_{i, \alpha}(l)} \right)$, it can also approximate $f(X)$ based on the triangle inequality:
\begin{equation}
   \sup_X \left|f(X)-\tilde{\mathbf{F}}(X)\right| \leq \delta.
\end{equation}
The number of parameters used to construct $\tilde{\mathbf{F}}$ is $NdLO(\ln(1/\delta))$.

\underline{Approximation by a quadratic network}: A quadratic neuron can precisely approximate a product function without incorporating extra error like a conventional network. To express $\tilde{\mathbf{F}}=\sum_{l=1}^L \widetilde{\times}_{i=1}^{N} \widetilde{\times}_{\alpha=1}^{d} X_{i, \alpha}^{\phi_{i, \alpha}(l)}$, the number of parameters used in a quadratic network is $4NdL$.
\end{proof}

\textbf{Remark 1.} The above constructive theorem informs us that a conventional network is hard to learn reflectionally invariant functions, \textit{i.e.}, much more parameters, but it is much easier for quadratic networks because quadratic neurons can easily represent a power term and a multiplication operation. Our main theorem heavily relies on \cite{han2022universal}. But our highlight is to permute the dimension, while their result permutes samples. Moreover, we intend to use quadratic neurons to reimplement this scheme to show their superiority in representing reflectional symmetry. One may ask if there are other kinds of constructs. Actually, one can also partition the domain into a lattice with a small grid size to approximate $f(X)$ instead of using the linear combination. But still, the quadratic network will show the advantages because the power term in the quadratic neuron is essentially suitable for reflectional invariance. 

\section{Cloud-RAIN}
Inspired by the above encouraging theoretical result, here we propose a point \underline{Cloud} method which enjoys the \underline{R}eflection\underline{A}l \underline{IN}variance (Cloud-RAIN) and whose spotlight is the incorporation of quadratic neurons. In addition, we also equip Cloud-RAIN with the PCA canonical transform, because the quadratic neuron alone can only tackle reflection across axis-aligned planes. When combining the PCA canonical transform, quadratic neurons can deal with reflection with respect to arbitrary planes.

\subsection{Quadratic Neurons}
Given a set of points $\mathbf{X}\in\mathbb{R}^{N\times C}$, where $N$ represents the number of points, and $F$ is the number of features. Usually, $C=3+c$ denotes $(x,y,z)$ position in the geometric space, and the feature number (\textit{e.g.}, $c=3$ for points only with color; $c=6$ for those with color and normals). Existing point cloud analysis methods usually focus on how to engineer $\mathbf{w}$ or $\f$ to fulfill the needs and characters of the problems. For example, RS-CNN~\cite{liu2019rscnn} restricts $\mathbf{w}$ to learn the relations between $p_i$ and $p_j$ in geometric space and extends a regular rigid CNN to the irregular configuration. DGCNN~\cite{wang2019dynamic} proposes the EdgeConv that operates on the edges between the central point and its neighbors and treats $\f$ as the edge features. Point Transformer \cite{zhao2021point} uses the vector self-attention operator to aggregate local features and the subtraction relation to generate $\mathbf{w}$. Despite these arts, they all use conventional neurons which integrate $\mathbf{w}$ and $\f$ in a linear fashion, as Eq. \eqref{eqn:linear} shows.

Here, the proposed Cloud-RAIN replaces the simple linear relation between $\f$ and $\mathbf{w}$ with a quadratic interaction by quadratic neurons. Theorem \ref{thm:representation} shows that quadratic neurons are powerful in learning a reflectionally invariant function.
        \vspace{-0.3cm}
\subsection{PCA Canonical Transform}
Note $(\x_1,\x_2,\cdots,\x_D)\to (\mathbf{\pi}_1\x_1,\mathbf{\pi}_2\x_2,\cdots,\mathbf{\pi}_D\x_D)$ is a special case of reflection symmetry, where the planes of symmetry align with the axes. Here, we introduce PCA canonical transform to empower the quadratic aggregation to deal with reflection across an arbitrary plane. We first show how the PCA transform is performed on the point cloud so as to obtain its canonical representation.  
Without loss of generality, let the point cloud be $\mathbf{X}\in\mathbb{R}^{N\times 3}$, $\mathbf{X}_{i} \in \mathbb{R}^{3\times 1}$  be the  $i$-th  point of  $\mathbf{X}$, and $\overline{\mathbf{X}}=\large(\sum_{i=1}^N \mathbf{X}_{i} \large)/N \in \mathbb{R}^{3\times 1}$ be the geometric center of  $\mathbf{X}$, we perform the PCA decomposition for $\mathbf{X}$ as follows:

\begin{equation}
    \frac{\sum_{i=1}^N \left(\mathbf{X}_{i}-\overline{\mathbf{X}}\right)\left(\mathbf{X}_{i}-\overline{\mathbf{X}}\right)^{T}}{N}=\mathbf{V} \boldsymbol{\Lambda} \mathbf{V}^\top,
    \label{eqn:PCA}
\end{equation}
where $\mathbf{V} \in \mathbb{R}^{3\times 3}$ is the matrix composed of eigenvectors ($\mathbf{v}_{1}, \mathbf{v}_{2}, \mathbf{v}_{3}$) and  $\boldsymbol{\Lambda}=   \operatorname{diag}\left(\lambda_{1}, \lambda_{2}, \lambda_{3}\right)$  are the corresponding eigenvalues. By multiplying the point cloud with the eigenvector matrix, we obtain the canonical representation of the point cloud $\mathbf{X}$ as $\mathbf{X}_{\text {can }}=\mathbf{XV}$.
    
The reflection-invariant property of  $\mathbf{X}_{\text {can }}$ is illustrated in the following: Applying a random reflection matrix  $\mathbf{F} \in \mathbb{R}^{3\times 3}$ on the point cloud  $\mathbf{X}$ to get a reflected point cloud $\mathbf{XF}^\top$ and performing PCA decomposition for $\mathbf{FX}$ as shown in Eq. \eqref{eqn:PCA}, we have the following:
\begin{equation}
\begin{aligned}
& \frac{\sum\left(\mathbf{F X}_{i}-\mathbf{F} \overline{\mathbf{X}}\right)\left(\mathbf{F X}_{i}-\mathbf{F} \overline{\mathbf{X}}\right)^\top}{N} \\
=& \mathbf{F}\left(\frac{\sum\left(\mathbf{X}_{i}-\overline{\mathbf{X}}\right)\left(\mathbf{X}_{i}-\overline{\mathbf{X}}\right)^\top}{N}\right) \mathbf{F}^\top\\
=&(\mathbf{F V}) \boldsymbol{\Lambda}(\mathbf{F V})^\top,
\end{aligned}
\end{equation}
where $\mathbf{FV}$ is the eigenvector matrix of the reflected point cloud $\mathbf{XF}^\top$. Therefore, the canonical representation $\mathbf{XF}^\top$ is 
$\left(\mathbf{X F}^{\top}\right)_{\text {can}}=\mathbf{X} \mathbf{F}^{T} \cdot \mathbf{F V}=\mathbf{XV}$. Thus, the effect of the reflection is counteracted in $\mathbf{X}_{\text {can }}$.

Although employing the canonical representation can realize the reflection invariance with respect to arbitrary planes, it brings up the issue of sign ambiguity. Given an eigenvector $\mathbf{v}$, both choosing $+\mathbf{v}$ and $-\mathbf{v}$ fulfills the rule of PCA transform. What's undesirable is we don't know if it is $+\mathbf{v}$ or $-\mathbf{v}$ being selected. Because $\mathbf{V}=[\mathbf{v}_1, \mathbf{v}_2, \mathbf{v}_3]$, there are eight possible combinations ($[\pm \mathbf{v}_1, \pm \mathbf{v}_2, \pm \mathbf{v}_3]$) that cause sign ambiguity in the canonical representation of $\mathbf{X}_{\text {can }}$. This is the intrinsic problem of using the canonical representation. However, quadratic aggregation can overcome the sign ambiguities due to its power term based on our earlier analyses. 

\textbf{Remark 2} (Theoretical guarantee). The canonical representation alone can deal with reflection across the arbitrary plane but suffers sign ambiguity, while the quadratic aggregation alone is invariant to the sign flipping but only works for reflection across axis-aligned planes. It is the combination between the canonical representation and quadratic aggregation that realizes the invariance to reflection across arbitrary planes. The advantage of such a combination is that it has a theoretical guarantee to be invariant to reflection, which will be confirmed by our experiments later. 
\begin{table}[!tb]
    \centering
    \small
    \scalebox{1.0}{
    \begin{tabular}{c|c|c|c|c}
       \toprule
    Dataset &  \multicolumn{2}{c|}{S3DIS}&  \multicolumn{2}{c}{ScanNet}\\
    \toprule
    Methods &  mAcc & mIOU &  mAcc & mIOU\\
    \midrule
    PointNet\cite{qi2017pointnet}  & 53.1 & 45.0 & 44.3 & 41.9\\
    PointNet\cite{qi2017pointnet}+Ours & 56.5 & 48.1 & 45.6 & 43.7\\
    PointNet++\cite{qi2017pointnet++} & 62.0 & 53.9 & 53.4 & 54.4\\
    PointNet++\cite{qi2017pointnet++}+Ours& 64.4 & 56.0 & 57.3 & 57.6 \\
    DGCNN\cite{wang2019dynamic}& 57.0 & 49.3 & 48.0 & 46.7  \\
    DGCNN\cite{wang2019dynamic}+Ours& 61.9 & 50.6 & 51.0 & 49.8\\
    LDGCNN\cite{zhang2019linked}& 59.1 & 49.9 & 50.6 & 49.9\\
    LDGCNN\cite{zhang2019linked}+Ours& 61.4 & 50.6 & 51.6 & 51.8\\
    GSNet\cite{xu2020geometry}& 52.7 & 44.5 & 43.7 & 45.2\\
    GSNet\cite{xu2020geometry}+Ours& 54.5 & 46.2 & 44.9 & 46.4\\
    ShellNet\cite{zhang-shellnet-iccv19}& 58.3 & 49.5 & 49.7 & 49.0\\
    ShellNet\cite{zhang-shellnet-iccv19}+Ours& 61.2 & 50.4 & 50.8 & 51.7\\
    PointMLP\cite{ma2022rethinking} & 80.3 & 70.1 & 68.9 & 69.3\\
    PointMLP\cite{ma2022rethinking} + Ours & 81.5 & 70.9 & 70.1 & 70.5\\
    PointMixer\cite{choe2022pointmixer} & 80.5 & 70.7 & 69.7 & 70.6\\
    PointMixer\cite{choe2022pointmixer} + Ours & 81.7 & 71.8 & 70.9 & 71.8 \\
    \bottomrule
    \end{tabular}}
    \vspace{-0.3cm}
    \caption{Experiments on S3DIS and ScanNet.}
    \vspace{-0.5cm}
    \label{tab:seg}
\end{table}
\begin{table*}[!tb]
    \centering
    \small
    \scalebox{1.0}{
    \begin{tabular}{c|c|c|c|c|c|c|c|c}
    \toprule
    \multirow{2}{*}{Model} & \multicolumn{2}{c|}{$x$ axis} & \multicolumn{2}{c|}{$y$ axis} & \multicolumn{2}{c|}{$z$ axis} & \multicolumn{2}{c}{$xyz$ axis} \\
    \cline{2-9}
    \rule{0pt}{10pt}
    &$\Delta$mAcc & $\Delta$mIOU & $\Delta$mAcc & $\Delta$mIOU & $\Delta$mAcc & $\Delta$mIOU & $\Delta$mAcc & $\Delta$mIOU\\
    \midrule
    PointNet \cite{qi2017pointnet} & $\downarrow3.5\%$ & $\downarrow4.1\%$ & $\downarrow2.6\%$ & $\downarrow3.1\%$ & $\downarrow45.0\%$ & $\downarrow43.8\%$ & $\downarrow45.8\%$ & $\downarrow44.2\%$\\
    PointNet \cite{qi2017pointnet} + Ours & $\downarrow0\%$ & $\downarrow0\%$ & $\downarrow0\%$ & $\downarrow0\%$ & $\downarrow0\%$ & $\downarrow0\%$ & $\downarrow0\%$ & $\downarrow0\%$\\
    \midrule
    DGCNN \cite{wang2019dynamic} & $\downarrow3.8\%$ & $\downarrow4.4\%$ & $\downarrow3.5\%$ & $\downarrow3.6\%$ & $\downarrow44.3\%$ & $\downarrow49.2\%$ & $\downarrow45.3\%$ & $\downarrow49.3\%$ \\
    DGCNN \cite{wang2019dynamic} + Ours & $\downarrow0\%$ & $\downarrow0\%$ & $\downarrow0\%$ & $\downarrow0\%$ & $\downarrow0\%$ & $\downarrow0\%$ & $\downarrow0\%$ & $\downarrow0\%$\\
    \midrule
    PointMLP \cite{ma2022rethinking} & $\downarrow4.3\%$ & $\downarrow5.1\%$ & $\downarrow3.9\%$ & $\downarrow4.1\%$ & $\downarrow64.7\%$ & $\downarrow68.9\%$ & $\downarrow66.1\%$ & $\downarrow69.2\%$ \\
    PointMLP \cite{ma2022rethinking} + Ours & $\downarrow0\%$ & $\downarrow0\%$ & $\downarrow0\%$ & $\downarrow0\%$ & $\downarrow0\%$ & $\downarrow0\%$ & $\downarrow0\%$ & $\downarrow0\%$\\
    \midrule
    PointMixer \cite{choe2022pointmixer} & $\downarrow4.9\%$ & $\downarrow5.7\%$ & $\downarrow4.4\%$ & $\downarrow4.9\%$ & $\downarrow65.1\%$ & $\downarrow69.4\%$ & $\downarrow66.5\%$ & $\downarrow69.9\%$ \\
    PointMixer \cite{choe2022pointmixer} + Ours & $\downarrow0\%$ & $\downarrow0\%$ & $\downarrow0\%$ & $\downarrow0\%$ & $\downarrow0\%$ & $\downarrow0\%$ & $\downarrow0\%$ & $\downarrow0\%$\\
    \bottomrule
    \end{tabular}}
    \vspace{-0.3cm}
    \caption{Performance drops on different reflection operations.}
    \label{tab: reflection}
    \vspace{-0.6cm}
\end{table*}
        \vspace{-0.3cm}

\section{Experiments}
To validate the reflection invariance of the proposed Cloud-RAIN, we conduct extensive experiments on point cloud semantic segmentation with multiple recent networks as a backbone. Other supplementary results such as model complexity analysis are put into the supplementary materials.
        \vspace{-0.5cm}
\subsection{Experimental Setups}
\textcolor{black}{We mainly use S3DIS \cite{armeni20163d} and ScanNet \cite{dai2017scannet}, the mainstream point cloud semantic segmentation benchmarks, to evaluate our proposed methods. We use the widely-adopted mean accuracy (mAcc) and IOU (mIOU) as evaluation metrics. For S3DIS, following the protocol of previous work \cite{wang2019dynamic, cui2021geometric, wang2021tree, qi2017pointnet, zhang-shellnet-iccv19, qi2017pointnet++}, we train for 100 epochs with 4 Tesla V100 GPUs. The batch size is set to 32. Following the standard practice, the raw input points are firstly grid-sampled to generate $4,096$ points. Unless otherwise specified, we use scale and jitter as data augmentation. For ScanNet, we train for 100 epochs with weight decay and batch size set to 0.1 and 32, respectively. The number of input points is set to be $8,192$ by sampling. Except for random jitter, the data augmentation of the ScanNet is the same as that of the S3DIS. We re-implement the original models for a fair comparison and only replace the linear aggregation with the quadratic one without changing other elements of the model. }
        \vspace{-0.3cm}
        
\subsection{Segmentation Results} We report the basic segmentation results in Table \ref{tab:seg}. All compared models are either classic models or recently published in flagship venues. The spotlight is that our proposed Cloud-RAIN can universally boost the existing models' performance by a reasonably large margin. In detail, integrated with our Cloud-RAIN, the current models' mAcc and mIOU scores are improved by at least $1.8\%$ and $0.7\%$ on the S3DIS benchmark, respectively. For a simple model like PointNet, our proposed Cloud-RAIN can boost the performance on mAcc and mIOU by $3.4\%$ and $3.1\%$. On the ScanNet benchmark, the improvement over ScanNet is not on par with that on the S3DIS dataset. However, our proposed methods can still escalate the mAcc and mIOU scores by at least $1.0\%$ relative to the existing methods. We think that if normal features over the ScanNet benchmark are considered, the improvement could be more significant. Although reflection symmetry is the primary goal of this manuscript, we would like to note to readers that the Cloud-RAIN is an effective drop-in replacement to escalate the performance of the original models.

\vspace{-0.4cm}
\subsection{Reflection Invariance Analysis}
Now, we provide detailed experiments to analyze the reflection-invariance properties of our proposed Cloud-RAIN. In the first part, we concentrate on reflections across axis-aligned planes that are very common in realistic scenarios such as symmetric street views. No PCA transform needs to be used because quadratic aggregation alone can deal with such reflection operations. In the second part, reflection across arbitrary planes is tested. 

\underline{Reflection invariance (axis-aligned planes).} To analyze the effect of data reflection, we select PointNet, DGCNN, PointMLP, and PointMixer as examples to conduct experiments on the S3DIS benchmark. In the experiment, we flip the point clouds along the $x$, $y$, $z$, and $xyz$ axis, respectively, and test the segmentation performance of the well-trained models with the reflected inputs, accordingly. During the training process, reflection augmentation is not applied for the original models and those integrated with our proposed methods. The results are listed in Table \ref{tab: reflection}.

As seen in Table \ref{tab: reflection}, our proposed Cloud-RAIN can keep the performance for the reflected inputs thanks to the effect of power terms. This is mathematically guaranteed because power terms return the same value for the flipped inputs. In contrast, the existing methods cannot withstand the input reflection, evidenced by the dramatic performance drop. Especially when the reflection is applied to the $z$ axis, the performance drop is devastating. This might be because the dynamic range of positions along the $z$ axis is far more extensive than that along the $x$ and $y$ axes. Thus, the destruction of flipping the $z$ axis to the model is severer. 

\begin{figure*}[!bt]
    \centering
    \includegraphics[width=0.9\linewidth]{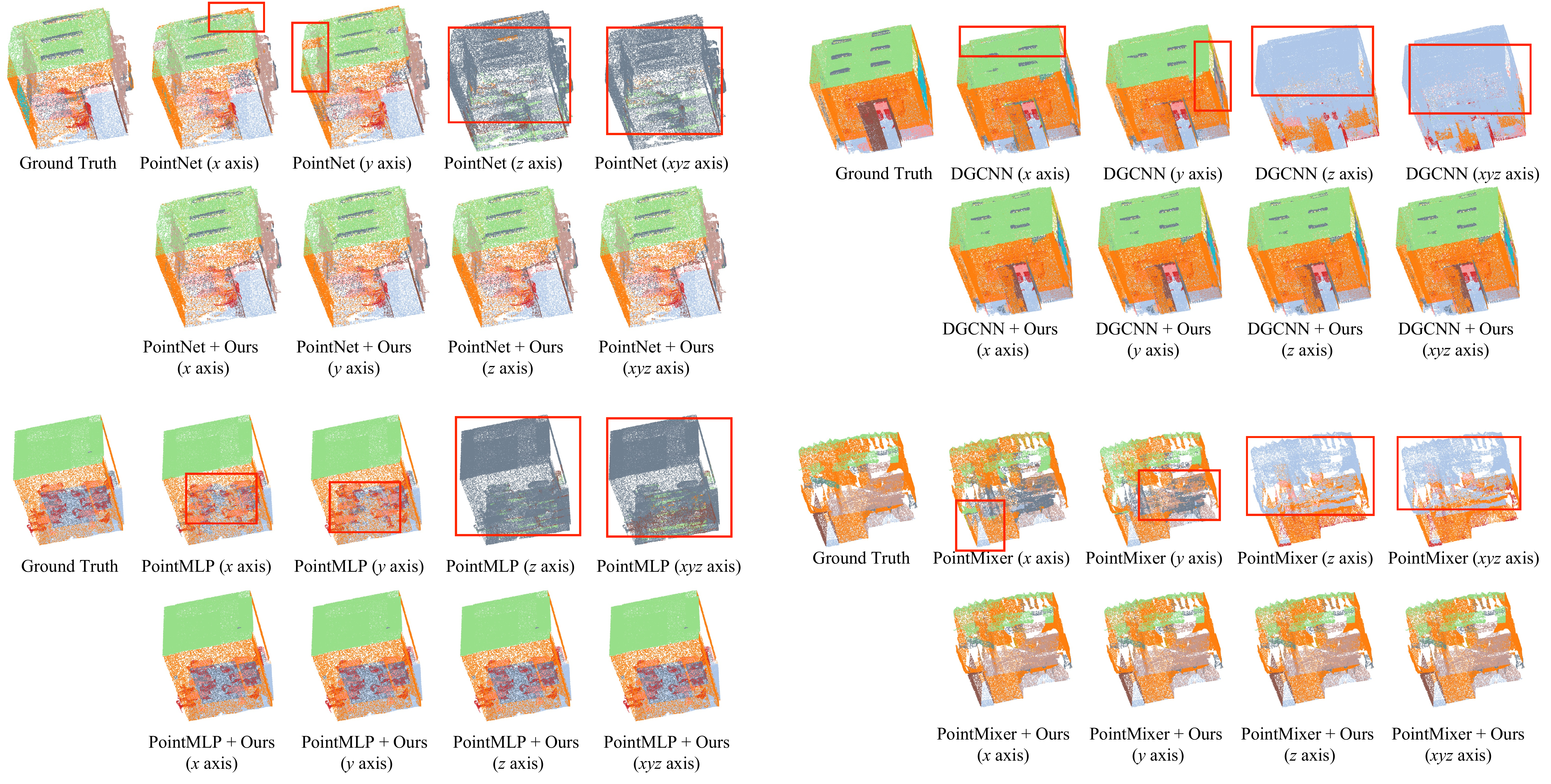}
    \caption{Comparison of the segmentation results with different reflections.}
    \vspace{-0.3cm}
    \label{fig: reflection}
        \vspace{-0.4cm}
\end{figure*}
\begin{figure}[!bt]
    \centering
    \includegraphics[width=0.8\linewidth]{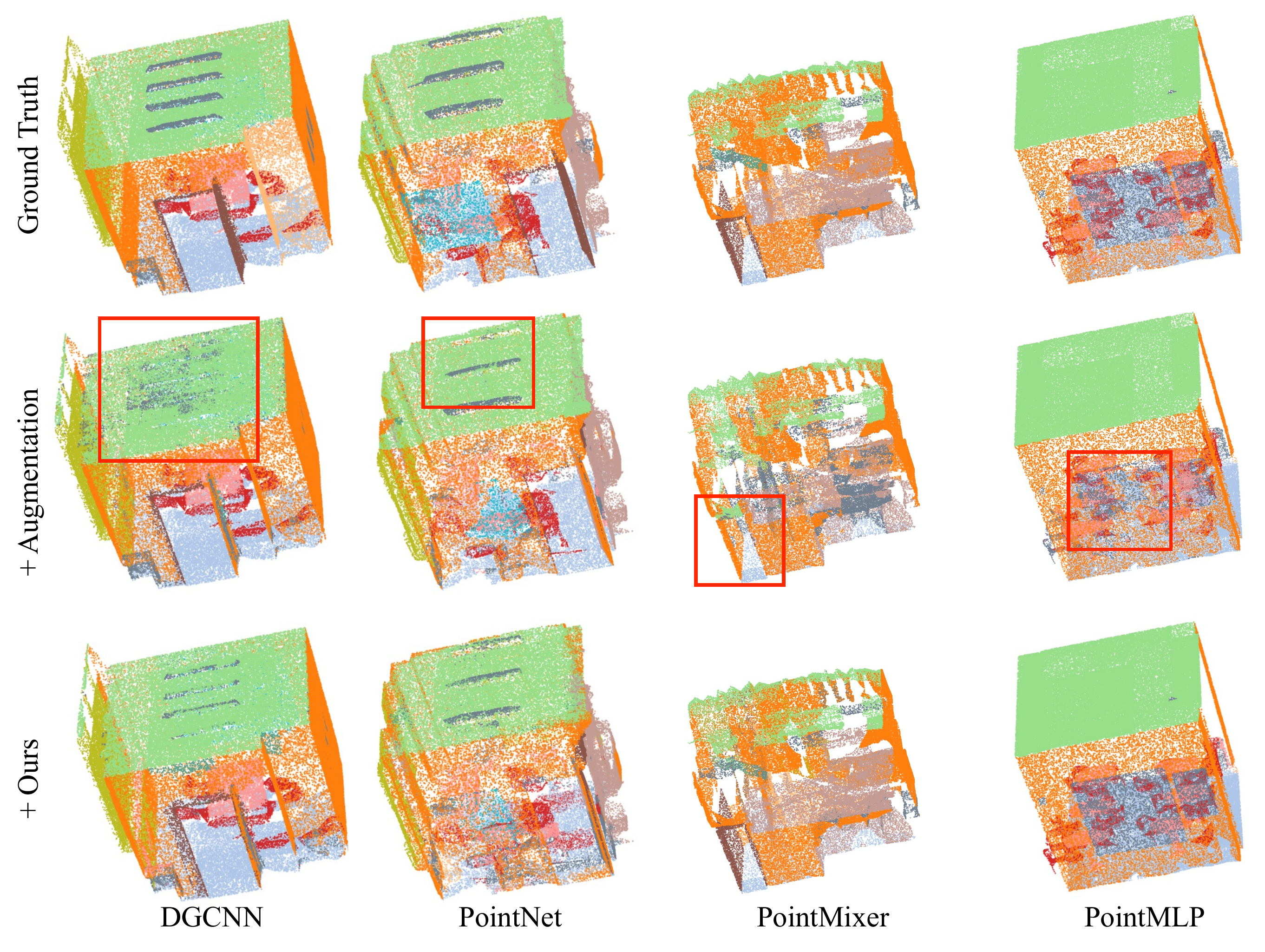}
            \vspace{-0.3cm}
    \caption{Comparison of the segmentation results from different models with reflection across randomly-selected planes.}
    \label{fig: rotateExample}
            \vspace{-0.7cm}
\end{figure}

Figure \ref{fig: reflection} offers semantic segmentation results in different reflection settings. For better visualization, we view the semantic segmentation results from the same angle regardless of their reflection. An interesting finding from Figure \ref{fig: reflection} is that when the point clouds are reflected along the $z$ axis, the original model is handicapped in distinguishing the categories of ceiling and floors. Many points with the ceiling category are misclassified into the floor class, indicating that the model just learns a reflectionally asymmetric mapping. 

\begin{table}[!bt]
    \centering
    \small
    \begin{tabular}{c|c|c}
    \toprule
    Model & $\Delta$mAcc & $\Delta$ mIOU \\
    \midrule
    PointNet & $\downarrow46.3\%(\pm 0.52\%)$ & $\downarrow42.7(\pm0.32\%)$\\
    PointNet + Aug & $\downarrow2.6\%(\pm0.46\%)$ & $\downarrow2.4\%(\pm0.28\%)$\\
    PointNet + PCA & $\downarrow1.7\%(\pm0.38\%)$ & $\downarrow1.9\%(\pm0.32\%)$\\
    PointNet + Ours & $\downarrow0\%$ & $\downarrow0\%$ \\
    \midrule
    DGCNN & $\downarrow45.9\%(\pm0.22\%)$ & $\downarrow40.4\%(\pm0.18\%)$\\
    DGCNN + Aug & $\downarrow8.0\%(\pm0.28\%)$ & $\downarrow8.5\%(\pm0.32\%)$\\
    DGCNN + PCA & $\downarrow6.8\%(\pm0.22\%)$ & $\downarrow7.7\%(\pm0.26\%)$\\
    DGCNN + Ours & $\downarrow0\%$ & $\downarrow0\%$\\
   \midrule
    PointMLP & $\downarrow66.9\%(\pm0.22\%)$ & $\downarrow69.8\%(\pm0.16\%)$\\
    PointMLP + Aug & $\downarrow10.2\%(\pm0.32\%)$ & $\downarrow11.3\%(\pm0.38\%)$\\
    PointMLP + PCA & $\downarrow7.4\%(\pm0.20\%)$ & $\downarrow8.6\%(\pm0.22\%)$\\
    PointMLP + Ours & $\downarrow0\%$ & $\downarrow0\%$\\
    \midrule
    PointMixer & $\downarrow67.1\%(\pm0.38\%)$ & $\downarrow70.0\%(\pm0.42\%)$\\
    PointMixer + Aug & $\downarrow11.0\%(\pm0.30\%)$ & $\downarrow11.9\%(\pm0.34\%)$\\
    PointMixer + PCA & $\downarrow8.6\%(\pm0.26\%)$ & $\downarrow9.4\%(\pm0.28\%)$\\
    PointMixer + Ours & $\downarrow0\%$ & $\downarrow0\%$\\
    \bottomrule
    \end{tabular}
            \vspace{-0.3cm}
    \caption{Performance of PointNet and DGCNN drops dramatically for reflection across arbitrary planes.}
    \label{tab: arbitRotate}
    \vspace{-0.5cm}
\end{table}

\underline{Reflection invariance (arbitrary planes).} We conduct experiments with PointNet, DGCNN, PointMLP, and PointMixer on the S3DIS benchmark to study the reflection across arbitrary planes. We randomly generate reflection planes, across which we flip the point clouds. To contrast the effectiveness of the proposed Cloud-RAIN, we apply data augmentation during the training process, where the training data are flipped along randomly generated planes. In the inference stage, we generate a plane randomly and flip the point clouds across the plane. For reliability of results, we run $5$ experiments and calculate the mean as well as the standard deviation. We also train the original models aided by PCA transform. All those comparative results are listed in Table \ref{tab: arbitRotate}.

From Table \ref{tab: arbitRotate}, we notice that without reflection data augmentation, all models suffer a severe performance loss for both mAcc and mIOU scores when there is no training augmentation. Even when the models are trained with randomly generated data augmentation, PointNet is still subjected to at least a $2.0\%$ drop on both evaluation metrics, while DGCNN's mAcc and mIOU scores drop at least $8.0\%$. What is favorable is that the performance of our models does not have any compromise with randomly generated reflection planes, which agrees with our theoretical analyses that Cloud-RAIN has the theoretical guarantee for reflection invariance. Furthermore, aided by PCA transform, the existing models are still subjected to the performance drop, which means these models cannot overcome the sign ambiguity given rise by PCA. This means that quadratic aggregation is necessary for reflectional symmetry over arbitrary planes.

We visualize and compare a representative example from the S3DIS with respect to an arbitrary plane reflection as Figure \ref{fig: rotateExample}. We sample $4,096$ points from the point clouds of each room and reflect those sampled points across the same randomly generated plane. Then, we merge the segmentation results to be the original room. From Figure \ref{fig: rotateExample}, we notice that when the test data are reflected across an arbitrary plane, the models trained without data augmentation cannot even accurately distinguish the categories like walls, ceilings, and so on. When it comes to the model trained with reflection data augmentation, though it is not severely affected by the arbitrary reflection during the inference time, there is still a structural distortion in distinguishing the boundaries compared with our proposed method, as highlighted in the red rectangle.

        \vspace{-0.2cm}
\section{Conclusion}
        \vspace{-0.1cm}

In this manuscript, we have identified the reflection invariance, alongside the rotation invariance, as a practical concern in point cloud applications. Then, we have derived a theorem suggesting that quadratic neurons can learn a reflectionally symmetric function much easier than conventional neurons do. Next, we have introduced quadratic neurons and PCA canonical transform to prototype a reflectionally-invariant model. Lastly, extensive comparison experiments show that embedding quadratic aggregation in the existing models can not only  strengthen the reflection invariance as our theoretical analyses illustrate. In the future, more efforts can be invested in exploring polynomial aggregation to unify rotational and reflectional invariance ultimately.
        \vspace{-0.3cm}
\bibliographystyle{ieeetr}
\bibliography{egbib}
\end{document}


\title{Supplementary Materials of ``Point Cloud Analysis for Reflectional Invariance"}

\author{
Yiming Cui$^{1}$,
Lecheng Ruan$^{2}$,
Hang-Cheng Dong$^{3}$,
Qiang Li$^4$,
Zhongming Wu$^5$,
Tieyong Zeng$^5$, 
Feng-Lei Fan$^{5*}$
\thanks{$^*$Feng-Lei Fan (hitfanfenglei@gmail.com) is the corresponding author.}
\thanks{
$^1$Yiming Cui (cuiyiming@ufl.edu) is with Department of Electrical and Computer Engineering, University of Florida, Gainesville, FL, USA.}
\thanks{$^2$Lecheng Ruan is with BIGAI, Beijing, China.}
\thanks{$^3$Hang-Cheng Dong is with School of Instrumentation Science and Engineering, Harbin Institute of Technology, Harbin, China.}
\thanks{$^4$Qiang Li is with Department of Electrical and Computer Engineering, McMaster University, Hamilton, ON, Canada. }
\thanks{$^5$Zhongming Wu, Tieyong Zeng, and Feng-Lei Fan are with Department of Mathematics, The Chinese University of Hong Kong, N. T. Hong Kong.}}


\maketitle

\IEEEpeerreviewmaketitle




\section{Model complexity.} 
Though our proposed method inevitably increases the number of parameters in the existing models, we argue that the reason for performance improvement by Cloud-RAIN is that the nonlinear aggregation can facilitate the extraction of useful features. To verify this argument, we increase the number of parameters in  PointNet \cite{qi2017pointnet} and DGCNN \cite{wang2019dynamic} by i) increasing the feature channels (`Wider') and ii) the number of layers (`Deeper'), and compare their semantic segmentation results with ours on the S3DIS benchmark in Table \ref{tab: complexity}. The results show that the proposed Cloud-RAIN still demonstrates a much better performance than the wider and deeper PointNet/DGCNN of more parameters, which validates the effectiveness of the nonlinear aggregation.

\begin{table}[hbt]
    \centering
    \scalebox{1.0}{
    \begin{tabular}{c|c|c|c}
    \toprule
    Method & $\#$ of params & mAcc & mIOU \\
    \midrule
    PointNet (Original) & 3.53M & 53.1 & 45.0 \\
    PointNet (Wider) & 3.95M & 53.4 & 45.2\\
    PointNet (Deeper) & 3.87M & 54.9 & 46.1\\
    PointNet+Ours & 3.81M & 56.5 & 48.1\\
    \midrule
    DGCNN (Original) & 0.98M & 57.0 & 49.3\\
    DGCNN (Wider) & 1.53M & 57.5 & 49.6\\
    DGCNN (Deeper) & 1.45M & 59.1 & 49.9\\
    DGCNN+Ours & 1.42M & 61.9 & 50.6\\
    \bottomrule
    \end{tabular}}
    \caption{The performance improvement by Cloud-RAIN is due to the nonlinear aggregation instead of the increased number of parameters. }
    \label{tab: complexity}
\end{table}

\bibliographystyle{ieeetr}
\bibliography{egbib}